*Research Article*

# Enhancing Robot Navigation Efficiency Using Cellular Automata with Active Cells

**Saleem Alzoubi[1] and Mahdi H. Miraz[2,3,4],***

[1]Jadara University, Irbid, Jordan
sa.alzoubi@jadara.edu.jo
[2]Xiamen University Malaysia, Sepang, Selangor, Malaysia
m.miraz@ieee.org
[3]Wrexham University, Wrexham, UK
m.miraz@ieee.org
[4]University of South Wales, Swansea, UK
m.miraz@ieee.org
*Correspondence: m.miraz@ieee.org

**Abstract:** Enhancing robot navigation efficiency is a crucial objective in modern robotics. Robots relying on external navigation systems are often susceptible to electromagnetic interference (EMI) and encounter environmental disturbances, resulting in orientation errors within their surroundings. Therefore, the study employed an internal navigation system to enhance robot navigation efficacy under interference conditions, based on the analysis of the internal parameters and the external signals. This article presents details of the robot's autonomous operation, which allows for setting the robot's trajectory using an embedded map. The robot's navigation process involves counting the number of wheel revolutions as well as adjusting wheel orientation after each straight path section. In this article, an autonomous robot navigation system has been presented that leverages an embedded control navigation map utilising cellular automata with active cells which can effectively navigate in an environment containing various types of obstacles. By analysing the neighbouring cells of the active cell, the cellular environment determines which cell should become active during the robot's next movement step. This approach ensures the robot's independence from external control inputs. Furthermore, the accuracy and speed of the robot's movement have been further enhanced using a hexagonal mosaic for navigation surface mapping. This concept of utilising on cellular automata with active cells has been extended to the navigation of a group of robots on a shared navigation surface, taking into account the intersections of the robots' trajectories over time. To achieve this, a distance control module has been used that records the travelled trajectories in terms of wheel turns and revolutions.

## 1. Introduction

Currently, robotisation of all branches of human activity is significantly expanding. Robots are replacing many tasks and functions which were previously performed by humans. Modern robots can be categorised into two large classes: stationary and mobile. Stationary robots perform various intelligent tasks exclusively within their designated locations. For instance, they can be permanently stationed on a conveyor belt, participating in the assembly of specific structure. In other words, a stationary robot is mounted on a stationary platform for performing its assigned tasks. In contrast, mobile robots are complex robotic systems affixed to moving platforms, with their motion precisely controlled by automation.

In third-generation mobile robots (intelligent robots), an important task involves enhancing robot navigation, enabling precise control the robot's movement within the given space, allowing it to reach specific points and perform assigned tasks. Key aspects of robot navigation include determining location, constructing a map of the surrounding environment, path planning and motion control.

Existing navigation systems rely on various approaches: inertial methods [1], GPS-based systems [2-6], LPS-based systems [7, 8] and simulation navigation systems [9-12]. In fact, all these approaches require a multitude of sensors, both internal and external signals as well as mobile communication systems. Consequently, the robot's navigation process become complex and highly dependent on external factors and surrounding environmental changes.

Approaches that use preliminary modelling of the robot's movement trajectory are worthy of attention [9-12]. These approaches do not rigidly tie the robot to the sensors responsible for movement. However, they are most effective for stationary environments where the surroundings remain relatively stable. The optimal solution to the navigation challenges lies in integrating all approaches to anticipate diverse scenarios that arise during navigation. In this context, our work addresses the problem of enhancing robot navigation using an embedded operational navigation map employing cellular automata with active cells.

## 2. Literature Review

There are four main categories of modern robot navigation systems:
- Inertial navigation systems (INS) [1];
- Global positioning system (GPS) [2-6];
- Local positioning system (LPS) [7, 8] and
- Simulated navigation systems [9-12].

Each of these systems possess unique characteristics and serves to address specific situational challenges. Each of these systems find widespread application across various industries, particularly where their effectiveness is most pronounced.

Inertial navigation systems (INS) are based on the use of inertial sensors and measuring instruments [13, 14]. Accelerometers and gyroscopes can be used as inertial sensors, using which, the robot can autonomously determine its location without relying on other navigation systems. However, autonomous use of INS is most effective for short distances with minimal obstacles. Combining INS with other navigation systems enhances overall effectiveness. In addition, the operation of such systems is significantly influenced by the topography of the surface on which the robot moves.

GPS uses satellite systems for navigation [2]. Robots exchange signals with satellites, enabling the satellite system to determine their location. The satellite signals provide the robots with necessary information about the next direction of movement. GPS based robot navigation systems excel in open areas with high navigation accuracy. However, it often fails in closed spaces, such as indoor rooms. In addition, expensive equipment is required to implement such GPS based system. In terms of security, unauthorised access may compromise the robot's navigation.

LPS are implemented based on the placement of local sensors, as well as various pointers in the form of signal beacons, which allow the robot to navigate in the local space [7]. These systems are typically used in enclosed spaces, such as automated warehouses, conveyors and restaurants. However, LPS based robot navigation systems have their own limitations: they make the robots dependent on the placement of various sensors and beacons, without which the robot cannot navigate effectively. In addition, LPS does not perform optimally in dynamically changing environments.

Simulation of navigation systems are very popular [9-12] and play a crucial role in testing and refining navigation algorithms and strategies without relying on physical robots. Such simulations can be implemented purely through software. For instance, a combined framework utilising CoppeliaSim and robot operating system (ROS) has gained popularity amongst the robot developers and researchers [9, 15]. Additionally, the robot virtual navigation (RVN) framework, based on OpenGL,[10] serves as one of the effective tools for simulating robot navigation. Another approach is to integrate the physical models with the software-based ones. For instance, the interactive robotics modelling & simulation System (IRMSS) [11], which consists of two subsystems: a physical model for robots and a 3D robotics modelling subsystem.

While virtual models dominate these systems, they often overlook the real-world unpredictability and the changes in the environment of the movement platform. Therefore, robotic experts are focusing on the autonomous navigation and path planning. One of highly researched algorithms in autonomous robot navigation is the use of graph theory, which plays a key role in finding optimal paths and determining efficient movement [16]. However, it requires pre-assessing the environment along the entire path, which can be challenging due to dynamic changes.

Heuristic path-finding algorithms using A* and D* also require preliminary terrain assessment and substantial computing resources [17–19]. A comprehensive overview of various methods and tools used for robot path planning is provided by Sánchez-Ibáñez *at al.* [20]. This paper [20] examines different types of cell shapes that define the robot movement spaces and discusses various interaction methods between robots and surfaces. Algorithms, such as local and global planning, reactive computing local optimisation, etc. were also evaluated, emphasising on their advantages and disadvantages. These algorithms use cell decompositions and roadmaps that are defined in advance. These algorithms require significant computing resources, and local optimisation may lead to local minima. Researchers continue to explore graph-based algorithms, which despite being time-consuming to construct, offer promising solutions.

There are numerous works that leverage cellular automata (CA) technologies for developing robot path planning algorithms [21–23]. The paper [21] presents a model, namely Genetic Shared Tabu Inverted Ant Cellular Automata (GSTIACA), for observing a group of robots by combining cellular automation technologies, genetic algorithms, ant algorithms and insights from the social behaviour of the pedestrians. The model initially implements a genetic algorithm, followed by CA technologies applied during specific navigation steps across diverse environments. The results demonstrate effective swarm behaviour of the robots. However, this model requires substantial computing resources and exhibits inertial since the individual robot behaviour model consumes time in analysing pheromones and queues, determining positions and also preliminary room identification. Additionally, the robot's analysis is confined to neighbouring cells within the Moore neighbourhood, without considering the broader environmental context of the movement platforms.

Syed and Kunwar [23] propose a robot path planning algorithm based on CA. The algorithm pre-processes the environmental contexts of the workspace employing image processing techniques using Minkowski sums. The CA generates a set of waypoints by analysing the activated parent and daughter cells within Moore neighbourhood. This algorithm is effective for short robot path distances. While the scale of the resulting image has a significant influence, it may not always be commensurate with the actual size of the robot. It's important to note that the algorithm pre-determines the optimal path, which, in real conditions, may not always be possible to implement. Furthermore, at each step of the movement, the robot estimates the path a single cell distance.

To address the potential existing issues, combining path planning with internal robot navigation allows the robot to adapt to unforeseen situations, minimising delays as well as avoiding unwanted trajectories. This combination allows the robot to function autonomously, continuously adjusting its movement trajectory even when deviations from the planned path occur, all while progressing toward its intended goal.

## 3. Navigation system for a robot working in a group

In robotics, a critical task involves coordinating the interaction of a group of robots during navigation and while performing tasks that require collective effort. To address these challenges of robot group navigation, several conditions must be met:
- All the robots working within the group must maintain appropriate distances from each other as well as from any obstacles and other moving objects.
- Each robot should be equipped with autonomous navigation system.
- The robots should possess the ability to identify the group member as well as objects within their field of view.
- The group must have the ability to jointly solve complex problems that cannot be tackled by a single robot.

To achieve these goals, each robot in the group must have both internal and external navigation capabilities. When navigating, the robots must detect the presence of obstacles, various stationary objects and moving entities in their paths. While stationary objects can often be bypassed, moving objects pose challenges. To address this issue, the robot's navigation system must predict the trajectory of such objects.

Additionally, robots within the same group maintain a minimum distance from each other, which can vary based on the assigned tasks.

To fulfil the aforementioned conditions, the robots also must have the necessary sensors. In fact, the robot's navigation system comprises three key components:
- Autonomous navigation unit;
- Sensor network; and
- Movement trajectory prediction unit.

While the autonomous navigation unit monitors the robot's travelled path, the sensor network facilitates interaction with the external environment. The motion trajectory prediction unit enables autonomous movement ensuring obstacle avoidance and collision prevention.

## 4. Functioning model of the robot control module

The robot control unit analogised with the black box model. To implement such a model, it is necessary to conduct a detailed analysis of the dynamics of the robot's behaviour. While the number of inputs corresponds to the number of sensors used, the number of outputs depend on the number of active mechanisms and their operating states. Interactions between these active mechanisms should also be taken into account, because these can significantly increase the number of states of the control unit.

It is also essential to comprehensively analyse all the potential situations that may arise during the task execution of the robots. The task execution period begins after receiving the task and ends upon completion. When multiple robots are involved, the task duration can vary.

Robots can navigate in two different modes, viz. offline navigation and task execution

The first mode characterises the complete autonomy of the robot's navigation and execution of functions, relying on the analysis of internal actions and calculations of various established quantitative values of the physical parameters of the robot elements. While in offline mode the robots evaluate the states of the internal elements instead of practically analysing the external signals, in task execution mode the robots mainly focus on analysis of the external signals coming from the sensor outputs.

In the autonomous operating mode of robot navigation, the initial states play a crucial role. These include:
- The robot's starting point on the location map;
- The target point(s) for navigation;
- The number of obstacles and their locations on the map;
- The robot's trajectory, accounting for obstacle avoidance as well as uneven terrain, such as ascents and descents.

After setting the target point, which can be specified in the form of conditional coordinates on the robot's navigation map, the computing unit calculates the potential short routes (if multiple trajectories exist) to reach that point. To achieve this, various algorithms can be used, such as graphs theories, ant algorithms, etc.

If the robot's navigation trajectory is divided into cells, then the distance is calculated using the following coordinates: $Y_d = Y_1 - Y_0$; and $X_d = X_1 - X_0$, where $Y_1$ and $Y_0$ are the vertical coordinates of the target point and, the robot location, respectively; whereas $X_1$ and $X_0$ are the horizontal ones. If $Y_0 = 0$, all necessary parameters are computed. Otherwise, (i.e., if $Y_0 \neq 0$), the nearest coordinate is considered, which determines the robot's turning point. The distance is then calculated using the following formula:

$$Y_d = Y_1 - (Y_{turn} - Y_0) \tag{1}$$

Where $Y_{turn}$ represents the vertical coordinate of the robot's closest turn.

After calculating the robot's trajectory to the specified point, the following calculations are performed:

The length of the wheel perimeter (L) is determined, which is typically known in advance.

The distance to the point with $Y_{turn}$ coordinates is taken into account and the number of wheel turns ($R_{Y_{turn}}$) of the robot is calculated. For example, if the distance to the turning point is 10 mitres (m), with a wheel length, $L_{wheel} = 1$ m, then the robot's wheels must complete 10 full revolutions in the forward direction on a flat surface.

The horizontal distance $X_d$ is then calculated and determine its sign to ascertain the rotation angle of the wheels. If the result is negative, then the robot's wheels turn to the left when it reaches the vertical coordinate $Y_{turn}$. If the surface is divided into rectangular cells, the rotation angles correspond to $90^0$ and movement occurs along these cells.

Next, the number of full and partial revolutions of the wheel is determined based on the algorithm.

The distance from the rotation coordinate to the coordinate of the next rotation point is also determined. Typically, these distances are calculated in advance before the robot begins navigation.

The robot's movement stops after all distances between turns have been covered and all turns have been completed. In practice, the following sequence of numbers (M) is stored in the robot's memory:

$$M = \langle (X_1, Y_1, Z_1, L_1), (X_2, Y_2, Z_2, L_2), \ldots, (X_N, Y_N, Z_N, L_N) \rangle \qquad (2)$$

Where $X_i, Y_i$ – indicate the coordinates of the i-th point or cell of rotation of the robot wheels; $Z_i = \{+90^0, -90^0\}$ – indicates the direction and angle of rotation on the cellular layout of the surface along which the robot moves; $L_i$- represents the distance (number of wheel revolutions) that the robot must travel from the (i-1)th turning point to the i-th turning point.

Having stored in the memory, the sequence allows the robot to sequentially transmits the coordinate values and the distances. It is to note that the distances are represented in wheel revolutions, not in meters. The number L may have a fractional part, indicating the amount of partial rotation of the wheel. After covering the distance to the next turning point, the robot's wheels turn by an angle determined by the Z value.

When the surface on which the robot moves has a complex terrain, more sophisticated methods are employed to calculate the distances between the turning points. These methods include: straight segments, arcs, integrals and others.

It is necessary to take into account the fact that in autonomous systems various types of errors have significant impacts on the accuracy of the task execution. Key error margins that affect the problem-solving precision include:

- Geometric shape of the wheel;
- Unevenness of the surface on which the robot moves;
- Errors in the accuracy of dividing the surface into cells;
- Inertia of rotation of the shaft on which the wheels are mounted.

To mitigate these error margins, the necessary confidence intervals for permissible error are established based on experimental studies.

The second mode operates the robot fully relying on the internal states of the environment in which the robot navigates. This mode is more reliable than the first one, as it also encompasses functionalities of the first mode.

Initially, the robot begins to move following the first mode. The initial direction of movement is calculated by finding the difference in the $X_d$ coordinate from the robot's starting point is calculated, using the formula $X_d = X_1 - X_0$. When $X_d = 0$, the robot moves solely along the Y coordinate. Otherwise (i.e., if $X_d \neq 0$), the sign of the difference is determined to ascertain the initial direction of movement. If $X_d < 0$, then the robot's wheels turn to the left, otherwise (i.e., if $X_d > 0$) the robot's wheels turn $90^0$ to the right.

Once the initial direction of movement is determined, the robot navigates by analysing the signals received as the inputs of the control unit from the outputs of various sensors. These sensors receive signals from special physical emitters, such as light and radio waves. In addition, these robots are also equipped with the sensors that react to objects located near it.

For normal robot movement, the control unit must have the following analysed inputs:

g0 – forward movement (logical "1");

g1 – movement to the left (logical "1"), movement to the right (logical "0");

g2 – presence of an obstacle on the right (logical "1"), absence of an obstacle on the right (logical "0");

g3 – presence of an obstacle on the left (logical "1"), absence of an obstacle on the left (logical "0");

g4 –presence of an obstacle in front (logical "1"), absence of an obstacle in front (logical "0");

g5 – presence of an obstacle behind (logical "1"), absence of an obstacle behind (logical "0");

g6 – stopping the robot (logical "1"); and

g7 – reverse movement (logical "1").

Input g1 receives signals from internal computers that calculate the direction of movement of the robot. Inputs g2, …, g5 receive signals from distance sensors that detect nearby objects. Typically, infrared or ultrasonic distance sensors are used.

The control unit generates control actions based on these aforementioned inputs.

Let us describe the functional purpose of each output of the control unit. In practice, the number of active states of the robot determines the number of outputs.

The potential states are as follows:
1. State of calm: No action occurs.
2. Forward motion state: The robot's wheels rotate in the forward direction.
3. Reverse motion state: The robot's wheels spin in the opposite direction.
4. Right turn state: The robot's wheels turn $90^0$ degrees to the right.
5. Left turn state: The robot's wheels turn $90^0$ degrees to the left.

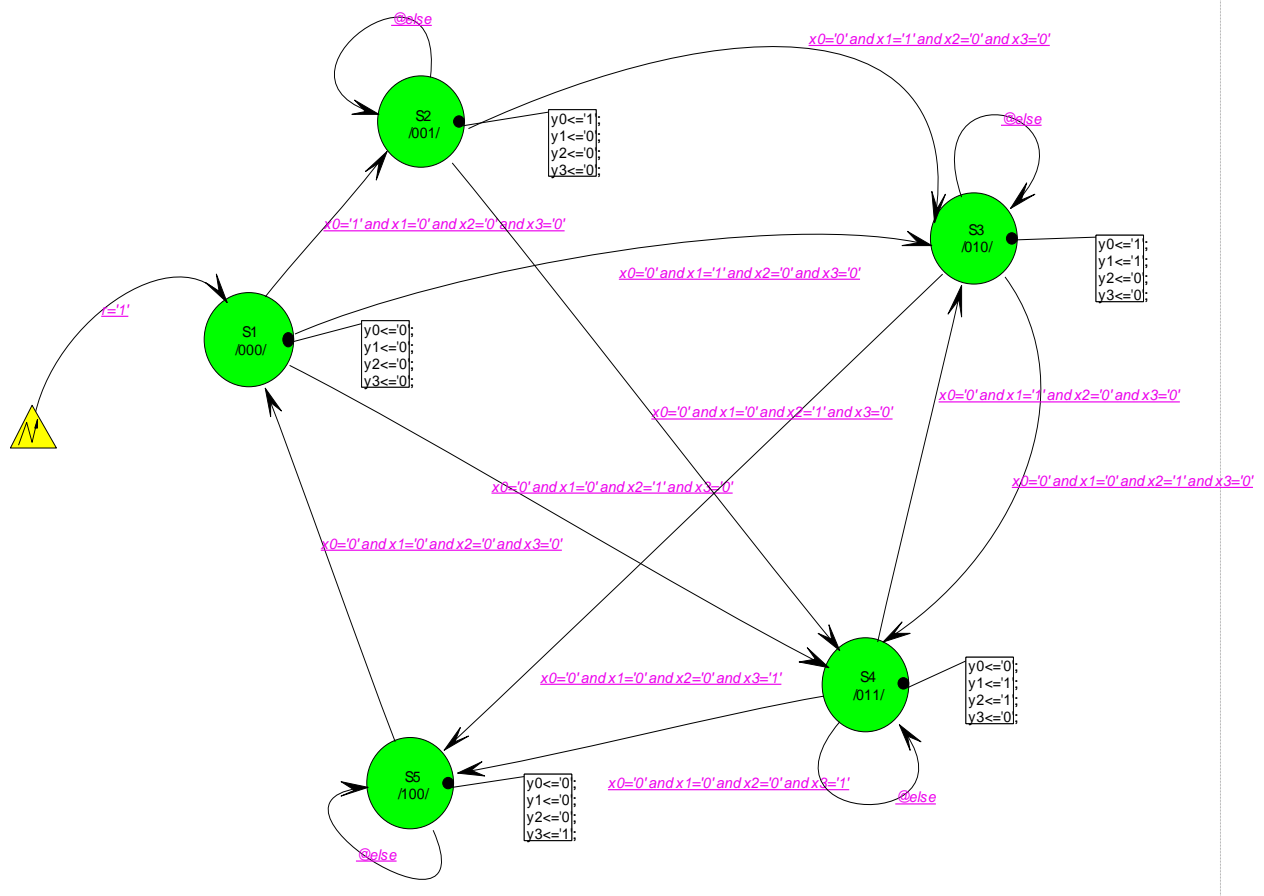

**Figure 1.** Graph of the control unit automaton obtained in the Active-HDL CAD system

Taking into account all possible states and input signals, we have compiled a sequence of actions in the form of an algorithm, for the operation of the control unit.
1. Initial state: All inputs contain logical "0" signals, except the zero output (q0=1), since this signal keeps all active mechanisms in a state of calm.
2. Upon receiving the coordinates of the target cell, a logical "1" signal appears at the zero input (g0=1) of the control unit. Consequently, a logical "1" signal is generated as the first output of the control unit (q1=1) and the robot moves forward.
3. While in motion, on-board sensors analyse signals that from different objects or obstacles. If the left sensor receives a signal (g1=1), a logical "1" signal is generated as the second output of the control unit (q2=1). The robot then turns its wheels $90^0$ to the left while continuing forward (q1=1).

4. Similarly, If the right sensor detects a signal (g2=1), a logical "1" signal is generated as the third output of the control unit (q3=1). The robot then turns its wheels $90^0$ to the right while continuing forward (q1=1).
5. When the sensor receives a signal (g3=1) indicating arrival at the destination cell, the robot stops (q0=0).
6. The movement ceases.

This algorithm can be presented using a truth table. Also, taking into account the number of inputs and outputs, the control unit's states can be described. Since four outputs are used, there could be 16 states. However, in this case, only six states are relevant. This robot navigation control unit is quite simple and can be implemented on an FPGA. An example of its implementation using the finite state machine editor is shown in Fig. 1.

**5. Robot navigation based on cellular technologies**

Previously discussed navigation methods allow robots to move with few obstacles but at the cost of reduced speed for reliability. The errors in navigation, as described before, over large distances from the target cell can lead to either significant deviations or delays due constant correction of the robot's trajectory or both.

To enhance navigation within obstacle-rich environments, the use of cellular technologies possesses great potentials [24 -27]. In such environments, traditional rules of cell transition cannot be applied since cell states are not considered in their classical representation. However, the fundamentals of cellular automata technologies with active cells can rather be used, which are described in [27, 28]. In this context, the active cell represents the robot's location at any given time.

The direction of movement depends on several parameters, such as:
Signs of the X and Y coordinates of the target cell where the robot aim to arrive; and
State of cells in the neighbourhood of the active cell in which the robot is currently located.

The first parameters, relating to the coordinates of the arrival cell at each moment of the robot's movement, are determined by the number of revolutions and turns of the robot's wheels. The second parameters determine the neighbouring cell, which will become active at the next step. That is, the robot will move into this cell. However, the robot cannot move to the robot cannot move to a neighbouring cell marked as an obstacle (logical "1"). Fig. 2 illustrates the coding of neighbourhood cells for moving the robot with $90^0$ turns.

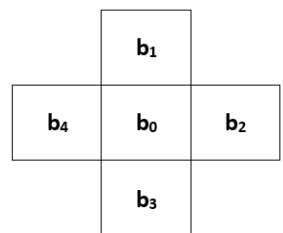

**Figure 2.** Coding of the neighbourhood cells for $90^0$ rotations

Cell b0 is the active cell, and cells b1, …, b4 constitute the von Neumann neighbourhood.
The algorithm for successful robot navigation is outlined below:
1. The navigation field is divided into rectangular cells (typically squares).
2. Determine the target cell where the robot aims to arrive.
3. Identify the initial cell where the robot is positioned.
4. Based on steps 2, 3, calculate the number of cells vertically and horizontally from the initial cell to the target cell.
5. Determine the direction of movement of the robot along the cells:
    a. If the difference $Y_d$ along the Y coordinate has a positive sign, the robot moves in the forward direction;
    b. If the difference $Y_d$ along the Y coordinate has a negative sign, the robot moves backward from the central cell (its initial location).

    c. If the difference $X_d$ along the Y coordinate of the target cell has a positive sign, the robot moves to the right, otherwise it moves left.
6. After determining the direction of movement, the robot begins to analyse groups of cells in a given direction that share a common cell. If one of the X or Y coordinates has a value of zero, the cells of the same group, located in the direction of movement, are analysed.
7. Based on the analysis, the robot moves to one of the cells within the analysed groups.
8. If the analysis shows that movement is impossible, then the robot analyses the remaining groups of cells until it finds a free cell.
9. If the analysis shows that the differences in the X and Y coordinates are both zero, the robot completes the movement; otherwise, it returns to step 4.

Analysis of groups of cells is carried out by simultaneous analysis of the nearest cells in the neighbourhood of the active cell as well as by analysis of the nearest cell from the cells in the neighbourhood. Initially, the direction of analysis of neighbourhood cells is established. The initial direction aims to reduce the vertical difference $Y_d$ between the coordinates of the initial and the target cells (vertical movement), followed by horizontal movement along the X coordinate.

At the start, cell $b_0$ analyses the first cell $b_1$ in its neighbourhood. If $b_1 = 0$, this cell then analyses its top cell of the neighbourhood, denoted as $b_1^1$. This can be described by the following formula:

$$b_1(t+1) = \begin{cases} 0, \text{if } b_0(t) = \text{active}, \ b_1^1(t) \wedge b_1^2(t) = 0 \\ 1, \ \text{in other cases} \end{cases} \quad (3)$$

To implement this approach, we will use the encoding shown in Fig. 3.

|   |   | $b_1^1$ |   |   |
|---|---|---|---|---|
|   | $b_{4,1}^{1,4}$ | $b_1$ | $b_{1,2}^{2,1}$ |   |
| $b_4^4$ | $b_4$ | $b_0$ | $b_2$ | $b_2^2$ |
|   | $b_{3,4}^{4,3}$ | $b_3$ | $b_{2,3}^{2,3}$ |   |
|   |   | $b_3^3$ |   |   |

**Figure 3.** Encoding neighbourhood cells for each cell

If cell $b_1(t+1) = 0$ at the next time step, the robot moves to this cell, and it becomes active. An example of the formation of a robot's movement trajectory for different obstacle locations is shown in Fig. 4.

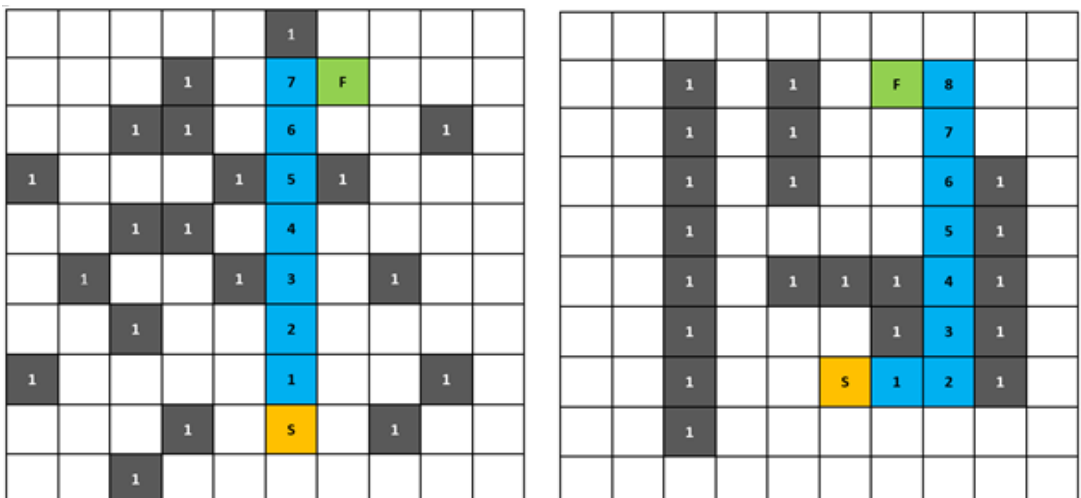

**Figure 4.** An example of the formation of robot's movement trajectory for different obstacle locations

As seen from the Fig. 4, obstacles are indicated by white logical "1" on a black cell background. The robot cannot enter these cells, and they remain inactive. Cells without obstacles are represented as empty and are coded as logical "0". A robot can enter these cells, making them active. Between the initial cell S and the target cell F, in both the examples, the numbers indicate the sequence of steps for the robot's movement. The number in each cell represents the time step during which the robot moves from one cell to another.

In the second example, the robot moved from cell S to cell F in eight time steps. The robot had to achieve zero horizontal difference. However, the robot to obtain a negative horizontal value by cell analysis since upward vertical movement was impossible due to the presence of obstacles.

Thus, cells in the neighbourhood of the active cell are analysed sequentially in a clockwise direction. If $b_1 = 1$, then cell $b_2(t)$ is analysed, and if $b_2(t)=0$, then its neighbourhood cells are analysed, according to the following expression:

$$b_2(t+1) = \begin{cases} 0, \text{if } b_0(t) = \text{active}, \quad b_{1,2}^{2,1}(t) \wedge b_2^2(t) = 0 \\ 1, \quad \text{in other cases} \end{cases} \quad (4)$$

Using the above scheme, the robot can move along a trajectory formed by the cellular environment. The robot's transition from one cell to another depends on the geometric dimensions of the cell relative to the robot's wheel size. Based on the number of wheel revolutions, the robot determines the transition to the next cell, using an internal wheel revolution counter.

The general algorithm for the functioning of a navigation cell within a cell area is presented in Fig. 5.

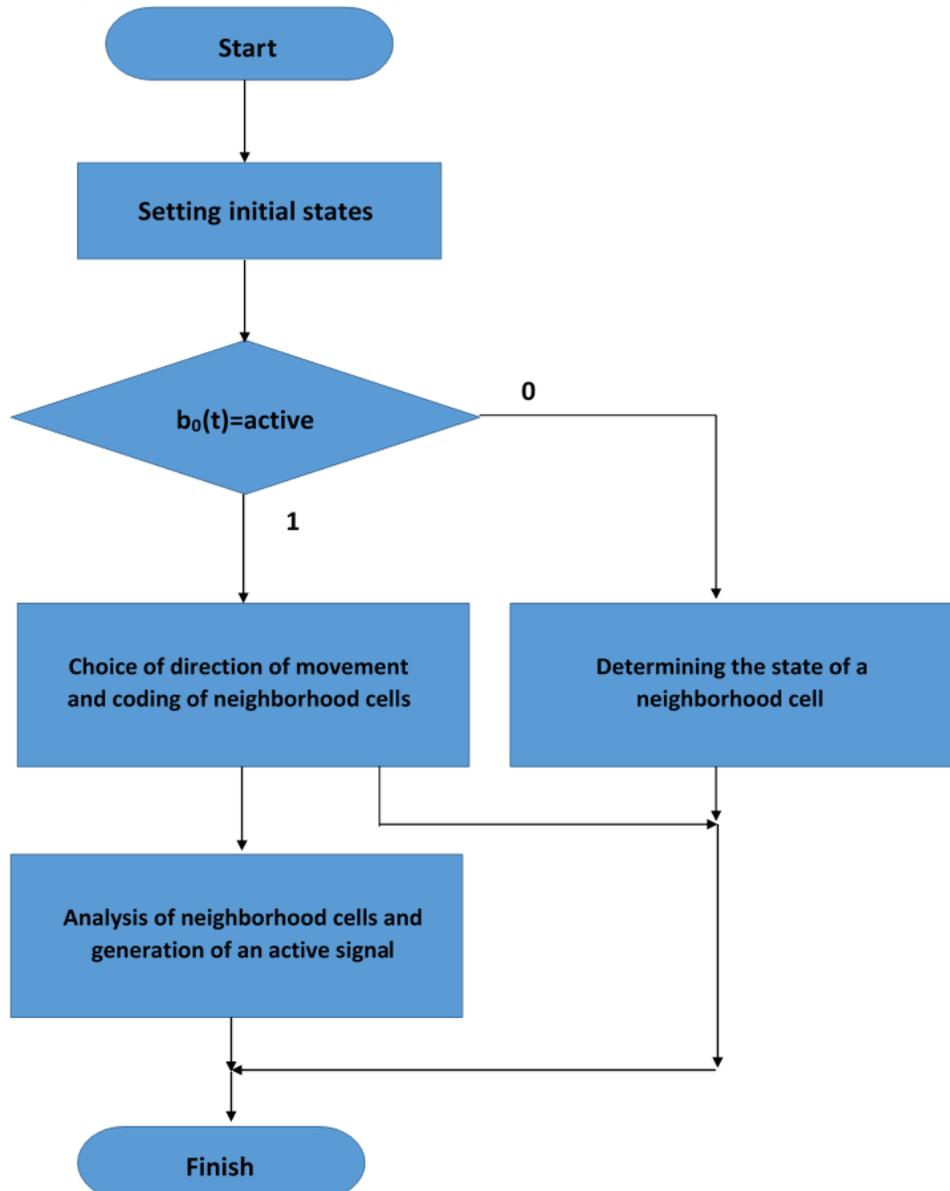

**Figure 5.** General algorithm for the functioning of a navigation cell within a cell area

According to the algorithm (Fig. 5), The cell assesses whether it is active (occupied by the robot). If the robot occupies a cell, that cell becomes active and selects the movement direction for the next time step. In accordance with this, the coding of the neighbourhood cells is determined. The algorithm that implements the choice of cell coding is presented in Fig. 6.

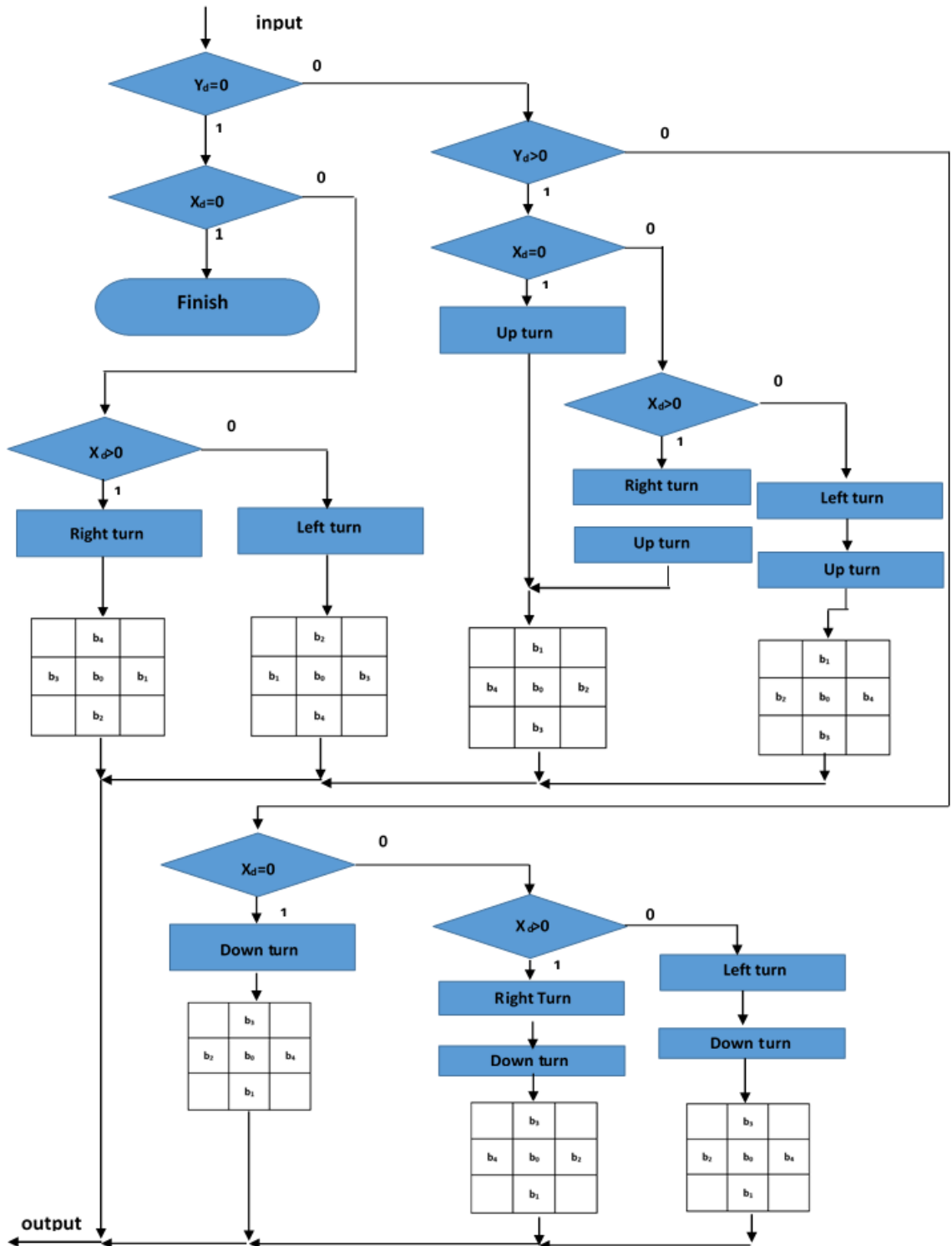

**Figure 6.** Algorithm for determining the direction of movement and coding neighbourhood cells

In the algorithm presented in Fig. 6, the values $X_d$ and $Y_d$ represent the differences between the X and Y coordinates of the initial and the target cells.

After determining the direction of movement, the cell analyses the neighbourhood cells and generates an active signal. The active signal generation algorithm is shown in Fig. 7.

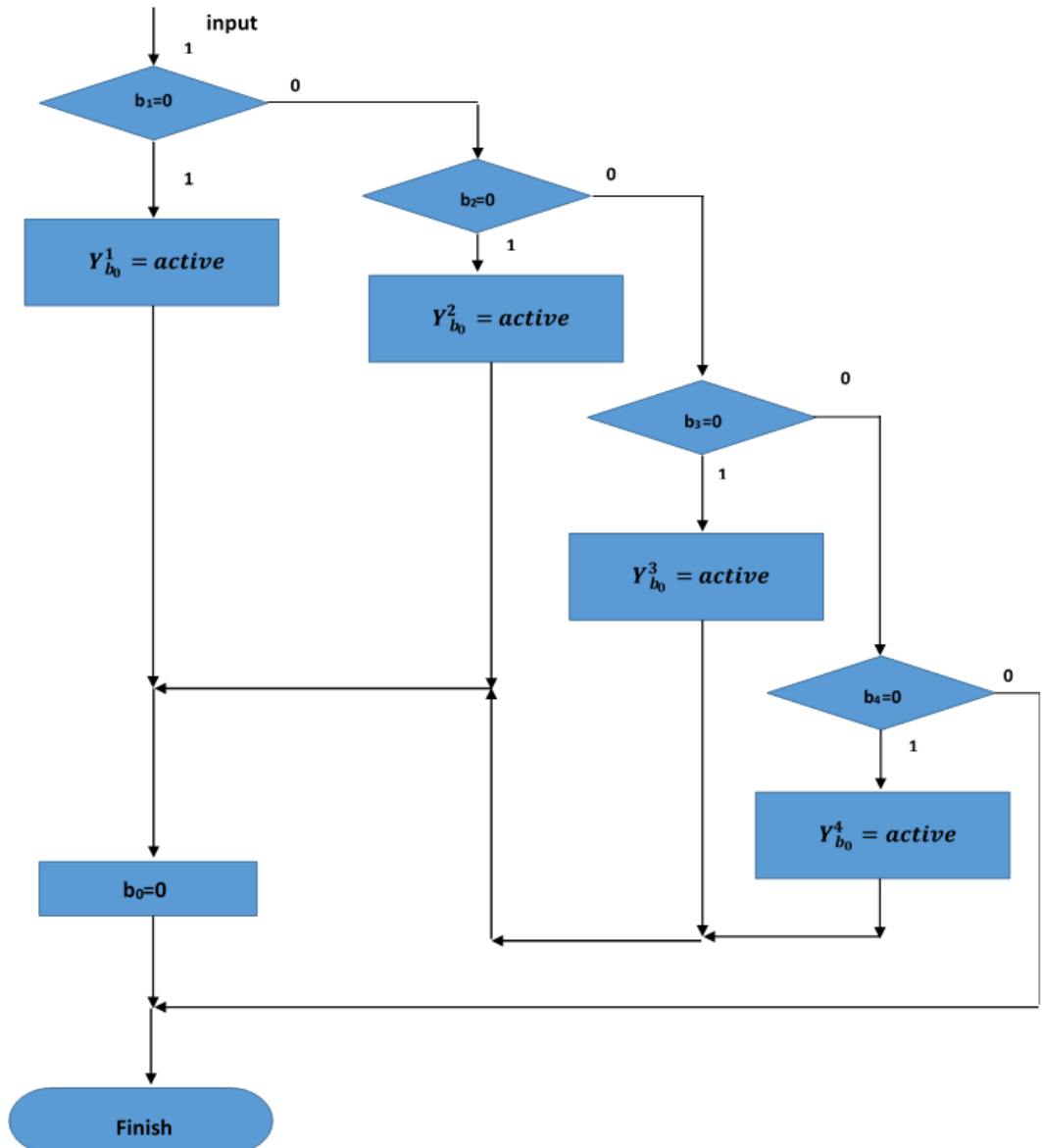

**Figure 7.** Block diagram of the active signal generation algorithm

According to this algorithm, as shown in Fig. 7, an active signal is generated at the active output that indicates the cell within the neighbourhood where the robot will move next. The active signal is determined by logical "1" at the corresponding i-th output $Y_{b_0}^i$, $(i = \overline{1,4})$.

If the cell is not active, then its state at the next time step is determined. The algorithm for an inactive cell state determination at the next time step is presented in Fig. 8.

The algorithm, as shown in Fig. 8, detects the presence of an active signal from one of the cells in its neighbourhood. In accordance with the neighbouring cell from which the active signal originates, the cells within the neighbourhood are encoded. Specifically, the cell opposite the one receiving the active signal is selected as the first cell of the neighbourhood. Once the encoding is formed, states are analysed in accordance with the selected direction of movement. To achieve this, input signals from neighbourhood cells are analysed in alignment with a specific direction of movement. For instance, the signal $X_{1,2}^a = 1$ indicates the direction of movement towards the cell in the neighbourhood of $b_1$ and $b_2$. In this case, the cell $b_1$ is analysed first, followed by $b_2$. In order for cell $b_0$ to assume the logical value "0" at time t+1, rules 136 and 172 according to Stephen Wolfram [16] must be fulfilled. The states of the cell according to rules 136 and 172 at the next time step are presented in Table 1.

**Table 1.** Implementation of rules 136 and 172 in elementary cellular automata

| $X_4(t)\ X_1(t)\ X_2(t)$ | 0 0 0 | 0 0 1 | 0 1 0 | 0 1 1 | 1 0 0 | 1 0 1 | 1 1 0 | 1 1 1 |
|---|---|---|---|---|---|---|---|---|
| Rule 136, $b_0(t+1)$ | 0 | 0 | 0 | 1 | 0 | 0 | 0 | 1 |
| Rule 172, $b_0(t+1)$ | 0 | 0 | 0 | 0 | 0 | 0 | 1 | 1 |

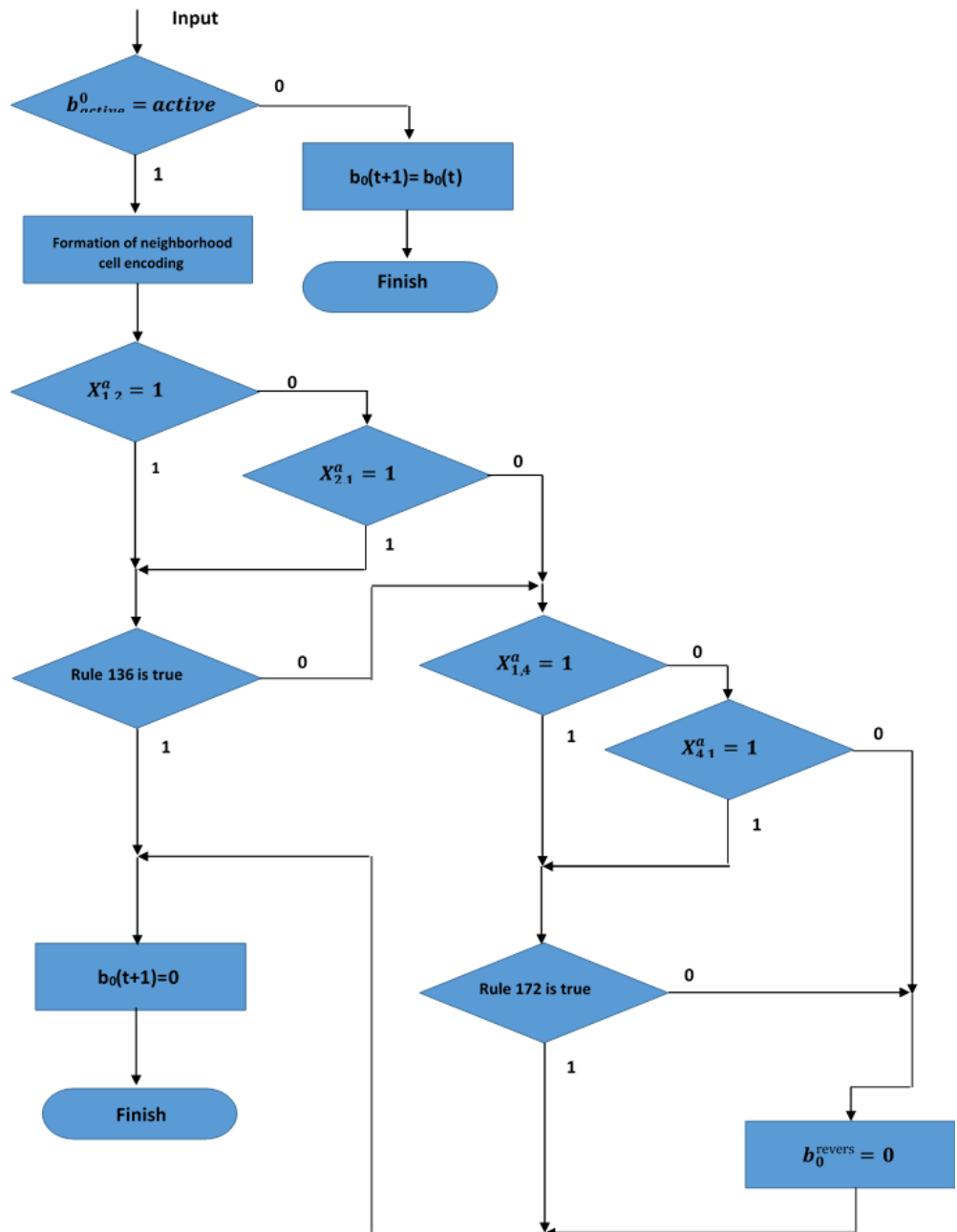

**Figure 8.** Algorithm for determination of the neighbourhood cell state at the next time step

Each robot contains an internal cellular environment that governs its navigation. The sizes of the cells can vary, but they must be large enough to accommodate the robot without extending beyond the cell boundaries.

The use of orthogonal mosaics significantly limits navigation, since it has a small number of movement options. Employing a hexagonal mosaic, as shown in Fig. 9, the navigation possibilities can be enhanced [25, 26].

The use of hexagonal mosaics increases the available of movement directions to six. The coding of neighbourhood cells is shown in Fig. 9 (b), where each cell shares a common side with one of the cells within the neighbourhood. The robot moves in the direction of one of the sides of the cell, in accordance with the calculated direction of movement. To determine the next cell the robot should move into, the three main cells of the neighbourhood aligned with the chosen direction of movement are analysed. If the direction of movement is towards cell $C_j$ ($j = \overline{1,6}$), the neighbouring cells considered are $C_{j-1}$, $C_j$, $C_{j+2}$. If j=1, then $C_{j-1}= C_6$ and if j=6, then $C_{j+1}= C_1$.

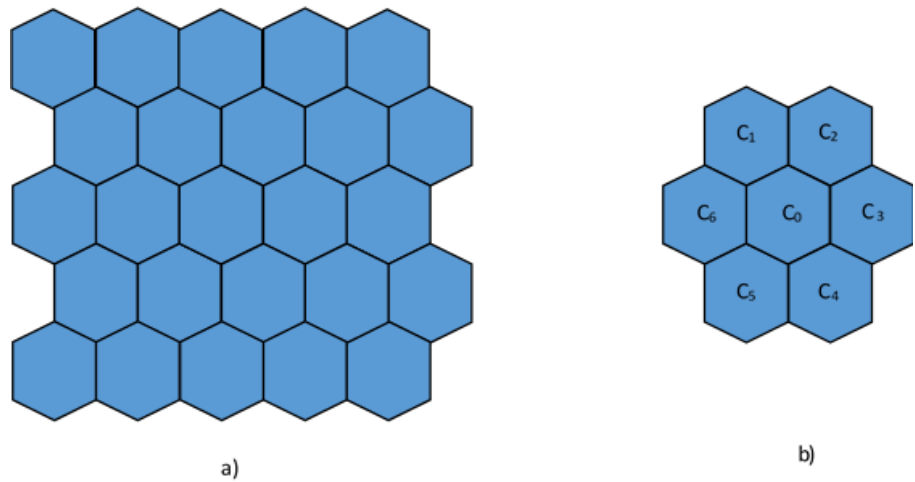

**Figure 9.** Hexagonal mosaic shape

If at least one of the three analysed cells in the neighbourhood has a logical "0" state, the control cell $C_0$= 0. This indicates that the robot can move to this cell and the cell will become active. The robot can move in the direction of increasing the difference between the coordinates of the current cell and the target cell, if and only if the condition $C_{j-1}(t) \wedge C_j(t) \wedge C_{j+1}(t) = 1$ is met. In such a case, the cell $C_{j+2}$ of the neighbourhood is analysed. This process applies for all the cells in the neighbourhood in a clockwise direction. If all the cells in the neighbourhood have a logical "1" state, the robot changes the direction of movement and, therefore, moves to another cell within the neighbourhood of its own active cell, which has smaller differences in coordinates.

## 6. Conclusion

The paper presented the principles of robot navigation from the starting point to a designated target. The principles combine insights from analysis of the internal characteristics as well as the environment of the movement surface. Leveraging cellular technologies based on cellular automata with active cells has significantly enhanced robot navigation. Notably, this method is especially effective when the movement map is formed in advance and all obstacles are already located within the cellular navigation environment. Remarkably, in such scenarios there is no need to use sensors.

Even when obstacles aren't predetermined, the method remains similarly efficient. By employing multiple distance sensors simultaneously scan the neighbouring cells, the robot acquires information about their states. By using the cellular environment, the robot then determines its next cell of movement. This whole process is completed within a single clock cycle, by taking advantage of the speed of the cellular automaton. Thus, this approach ensures highly reliable and efficient robot navigation.

In practice, the robot executes commands exclusively from the embedded navigation cell environment. The use of cellular automata significantly reduces the number of computational operations, allowing the navigation module to be implemented in the form of a finite state machine with hard logic. This design enhances the performance of the system, which relies on the switching time of the electronic gates.

Furthermore, based on cellular automata with active cells, group navigation of robots on a shared surface becomes straightforward, taking into account the intersections of their trajectories over time. To facilitate this approach, a distance control module keeps records of the trajectories of the travelled paths, in terms of the number of wheel turns and wheel revolutions.

Future research aims to explore navigation principles for group of robots using cellular automata with multiple active cells, facilitating collaborative tasks. The use of multiple active cells will help the group robots to avoid collisions amongst themselves. In such group navigation approach, each active cell will describe the trajectory of an individual robot in real time.